\documentclass[runningheads]{llncs}
\pdfoutput=1
\usepackage{url}  
\usepackage{graphicx}  

\usepackage{amssymb}
\usepackage{amsmath}
\usepackage{booktabs}
\usepackage{tikz}
\usepackage{rotating}
\usepackage{multirow}

\newcommand{\cons}[1]{\texttt{#1}}

\definecolor{red1}{rgb}{1.0, 0.5, 0.5}
\definecolor{red2}{rgb}{1.0, 0.7, 0.7}
\definecolor{red3}{rgb}{1.0, 0.9, 0.9}

\definecolor{black}{rgb}{0.0, 0.0, 0.0}
\definecolor{grey1}{rgb}{0.7, 0.7, 0.7}
\definecolor{grey2}{rgb}{0.9, 0.9, 0.9}

\definecolor{purple1}{rgb}{0.6, 0.0, 0.8}
\definecolor{purple2}{rgb}{0.91, 0.5, 1.0}
\definecolor{purple3}{rgb}{0.98, 0.9, 1.0}

\definecolor{blue1}{rgb}{0.0, 0.35, 0.7}
\definecolor{blue2}{rgb}{0.6, 0.79, 1.0}
\definecolor{blue3}{rgb}{0.9, 0.95, 1.0}

\newcommand{\best}[1]{\colorbox{blue1}{\color{white} #1}}
\newcommand{\almostbest}[1]{\colorbox{blue2}{#1}}
\newcommand{\closetobest}[1]{\colorbox{blue3}{#1}}

\newcommand{\whitebox}[1]{\colorbox{white}{#1}}

\begin{document}

\title{State Representation and Polyomino Placement for the Game Patchwork}
\author{Mikael Zayenz Lagerkvist\inst{1}\orcidID{0000-0003-2451-4834}}
\authorrunning{Mikael Z. Lagerkvist}
\institute{\email{research@zayenz.se}\\
\url{https://zayenz.se}}

\maketitle

\begin{abstract}
  Modern board games are a rich source of entertainment for many
  people, but also contain interesting and challenging structures for
  game playing research and implementing game playing agents.
  
  This paper studies the game Patchwork, a two player strategy game using
  polyomino tile drafting and placement. The core polyomino placement
  mechanic is implemented in a constraint model using \cons{regular}
  constraints, extending and improving the model
  in~\cite{Lagerkvist2008} with: explicit rotation handling; optional
  placements; and new constraints for resource usage.

  Crucial for implementing good game playing agents is to have great
  heuristics for guiding the search when faced with large branching
  factors. This paper divides placing tiles into two parts: a \emph{policy}
  used for placing parts and an \emph{evaluation} used to select among
  different placements. Policies are designed based on classical
  packing literature as well as common standard constraint programming
  heuristics. For evaluation, \emph{global propagation guided regret}
  is introduced, choosing placements based on not ruling out later
  placements.

  Extensive evaluations are performed, showing the importance of using
  a good evaluation and that the proposed \emph{global propagation
    guided regret} is a very effective guide.
\end{abstract}

\section{Introduction}
\label{sec:introduction}

Game playing has long been a core area of traditional AI
research. While classical board games such as Chess and Go are
computationally very hard, they are also somewhat simple in their
actual game play mechanics and representations. Modern board games,
such as Settlers of Catan, Ticket to Ride, Agricola, etc. are a rich
source of entertainment and creativity that often include complicated
game states and complex rules. The more complicated game states,
rules, and interactions represent a challenge for implementing game
playing logic. In particular, the branching factor is often very
large, and at the same time the complicated states make each step much
more computationally expensive.

Patchwork~\cite{game:patchwork} by Uwe Rosenberg is a two player
strategy game that uses polyomino tile drafting and placement. A
polyomino is a geometric form formed by joining one or more
squares edge to edge, for example as in Tetris. It was
released in 2014, and is in the top 100 games on Board Game
Geek~\cite{bgg:toplist}. The game is simple to describe for a human,
but the central tile placement mechanic is non-trivial to implement in
an effective and correct manner. This paper shows how to 
implement the tile placement sub-problem using constraint programming.

Game playing AI typically uses a tree-based search such as Minimax,
Alpha-Beta, or Monte-Carlo Tree Search. For reaching good performance,
it is crucial to have heuristics for guiding the tree search into
promising parts of the search tree quickly, AlphaGo~\cite{Silver2016}
is a prime example of this. In Patchwork, this requires having good
heuristics for packing polyominoes. This paper defines a
\emph{strategy} as a combination of a \emph{policy} used for creating
placements of the tiles and an \emph{evaluation} that chooses among
the generated placements.

Policies are designed based on classical packing literature, in
particular the Bottom-Left strategy from~\cite{Baker1980}.  In
addition and as contrast, standard constraint programming heuristics
are also used for placement policies, both simple heuristics such
as first-fail as well as more advanced modern heuristics such as
accumulated failure count/weighted degree~\cite{Boussemart04}. For
evaluation, \emph{global propagation guided regret} is introduced,
essentially choosing placements based on not ruling out later
placements. This is contrasted with more obvious measures such as
packing to the left or towards a corner.

\paragraph{Contributions} This paper proposes using constraint
programming for implementing parts of complex game states in modern
board games. It extends and improves a previous polyomino placement
model using \cons{regular} constraints, including optional placements
and explicit handling of transformations. A formulation for exact resource
usage for tiles using \cons{regular} constraints is also
introduced. For guiding search, a new and straight-forward heuristic
called \emph{propagation guided global regret} is designed, that uses
look-ahead and the results of propagation to guide search. Extensive
evaluation of the model and proposed heuristics is done.

\paragraph{Plan of paper} In the next section, some background on
constraint programming and AI for game play is given. In
Sect.~\ref{sec:patchwork} the game Patchwork is described in
detail. Sect.~\ref{sec:model} introduces the model for the core
placement problem, and the following section
introduces the placement heuristics
developed. The heuristics are evaluated in
Sect.~\ref{sec:experiments}, and finally some conclusions and
directions for future work are given.

\section{Background}
\label{sec:background}

This paper is concerned with implementing a representation of a part
of a game state using constraint programming. To give the context,
some general background on both game playing AI and constraint programming is
needed. 

\subsection{Game playing}
\label{sec:game-playing}

Game playing is a branch of AI where agents interact following a
certain set of rules. Common board games are typically
\emph{discrete} and \emph{sequential}. The number of potential actions
in each step is the \emph{branching factor} ($b$), and the
number of actions taken by each agent during a game is the number of
\emph{plies} ($d$). The full tree defined by the potential actions is
typically very large, in the order of $O(b^d)$.

Game playing is most often implemented using heuristic state space
exploring game tree search. Typical classical examples are Minimax and
Alpha-beta pruning. More recently, Monte-Carlo Tree Search (MCTS) has become
very influential. For a survey of MCTS methods and results,
see~\cite{Browne2012}. 

A core issue in implementing a game playing AI system is to represent
the game state. Key requirements are \emph{correctness}, \emph{speed},
and \emph{memory size}. Classical games such as Chess, Go, and
Othello/Reversi have fairly simple game states, and much effort has
been in creating very small and efficient representations. Modern
board games in contrast have more complex state spaces and rules,
which complicate game state implementation. Some recent examples of
implementing game AI for modern board games include Settlers of
Catan~\cite{Szita2009},
Scotland Yard~\cite{Nijssen2012},
7~Wonders~\cite{Robilliard2014},
and Kingdomino~\cite{Gedda2018}.

\subsection{Constraint programming}
\label{sec:cp}

Constraint programming is a method for modeling and solving
combinatorial (optimization) problems. Modeling problems with
constraint programming is done by defining the variables of the
problem and the relations, called constraints, that must hold between
these variables for them to represent a solution. A key feature is
that variables have finite domains of possible values.

Constraints can be simple logical and arithmetic constraints as well
as complicated global or structural constraints. Of particular
interest for this paper is the the \cons{regular} constraint
introduced by Pesant~\cite{Pesant2004}, where a specification for a
regular language (a regular expression or a finite automaton) is used
to constraint a sequence of variables.

\section{Patchwork}
\label{sec:patchwork}

Patchwork~\cite{game:patchwork} is the first game in a series of games
by Uwe Rosenberg that uses placing polyominoes as a core game
mechanic. Patchwork is the simplest of these games, with the polyomino
placement being front and center to the game play. It is a top-ranking
game on the Board Game Geek website, at place 64 out of more than a
hundred thousand entries~\cite{bgg:toplist} in July 2019, around five
years after the original release.

\begin{figure}
  \centering
  \includegraphics[width=0.8\textwidth]{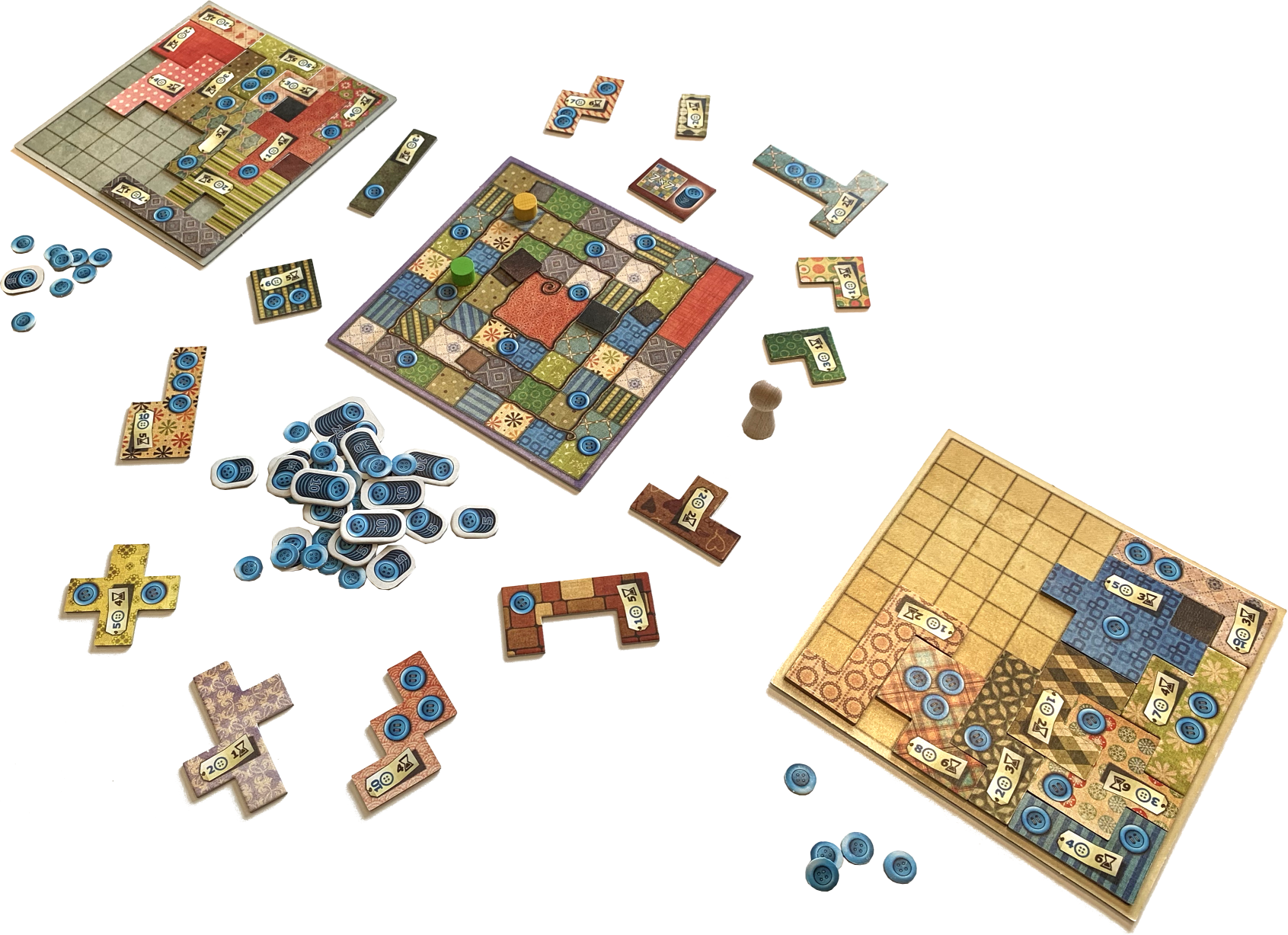}
  \caption{Patchwork game in progress}
  \label{fig:game-in-progress}
\end{figure}

\subsection{Rules}
\label{sec:patchwork:rules}

Patchwork is a two-player game with perfect information. The game
consists of 33 polyominoes called \emph{patches}, a marker,
currency markers called \emph{buttons}, two player tokens, two 9 by 9
boards (one per player) where patches are to be placed, and a central
time board. Each patch has a button cost, a time cost, and between 0
and 3 buttons for income. The time board has 53 steps in total. Five
of the steps have a special 1 by 1 patch, and 9 steps are marked for income.

At the start of the game, the patches are organized in a circle, with
a marker after the smallest patch. The player tokens are at the start
of the time board, and each player has an empty board and 5
buttons. The game is scored based on the number of buttons gained and
squares covered for each player at the end of the game.

In each turn, the player whose marker is the furthest back gets to
move (if both players are at the same place, the last to arrive there
gets to play). When making a move, a player may either advance their
time marker to the step after the other player or buy and place a
patch. Advancing the time marker to the step after the opponents time
token is always possible, and gives the number of steps in button
income.

To buy and place a patch, the player may choose one of the three next
patches in the circle. The player must pay the indicated number of
buttons on the patch (between 0 and 10) and place the patch on their
board. The marker is moved to the place of the bought patch, and the
player token is moved the number of steps indicated on the patch
(between 1 and 6). To buy a patch, the player must have sufficient
funds and the ability to place the patch on their board.

If the player when advancing passes a step marked for income on the
time board, they collect new buttons based on the number of income
buttons on the patches they have placed. If the player is the first to
pass one of the special 1 by 1 patches on the time board, they get the
patch and place it on their board immediately. 

The game ends when both players have reached the center of the time
board. The final score is the buttons they have acquired, minus two
for each uncovered square on the board. The first player (if any) that
filled a complete 7 by 7 area of their board gets an extra 7
points. For a full description of the rules,
see~\cite{game:patchwork}.

\subsection{Strategy}
\label{sec:patchwork:strategy}

The game requires balancing income, filling squares, and placing
patches to not disallow future patch placements. The time remaining
for a player can be viewed as a resource that is spent when making
moves.

Initially the player has $5 - 2\cdot9\cdot9=-157$ points. If no
patches are purchased (only advancing the time marker), the final
score for the player would be $-157 + 53=-104$.  A normal good score
for patchwork is positive, leading to a goal of earning at least 3
points per square advanced ($157/53\approx 2.96$).

Each patch can be evaluated in isolation for the total change in score
that patch would give. There are four main things to consider for a patch $P$:
\begin{itemize}
\item Size $S_P$, the number of squares the patch will cover.
\item Button cost $C_P$, the number of buttons to pay for the patch.
\item Time cost $T_P$, the number of steps to advance the time marker.
\item Button income $B_P$, the number of buttons to collect at income spots.
\end{itemize}
The total income gained by the end of the game when buying patch $P$
at time $t$ is determined by the number of remaining button income
spots on the board $I(t)$ (starting at 9 and mostly evenly spaced out).

Assuming that we are interested in maximizing our point gain per time
used for patch $P$ at time $t$, the following formula can be used
\begin{equation}
  \label{eq:gain}
G(P,t) = \frac{2\cdot S_P - C_P + I(t)\cdot B_P}{\min(T_P, 53-t)}
\end{equation}

Two additional issues relevant to the above evaluation are the 7-by-7
bonus and the size 1 squares on the board. The bonus is a property
that needs to be planned for, so we leave it to the planning. The 5
size 1 squares are each on their own worth 2 points and can easily be
added when evaluating $G$, but depend on the current game state. Their
main use however is in either completing a 7-by-7 square or
\emph{preventing} the opponent from completing their own 7-by-7
square. Again, we leave this out of the basic evaluation since it is
question of planning and not static evaluation.

\subsection{Characteristics}
\label{sec:patchwork:characteristics}

The game is as mentioned a 2-player game with
perfect information. The game set-up includes a shuffle of the
polyominoes giving $32!\approx 2.6\cdot 10^{35}$ different games. 

The branching factor can be quite large. The first part is the choice
of either advancing the time marker or buying one of the three next
patches (if possible). Given that a patch is bought, it must be placed
on the board. For example, in the initial step, a 2 by 3 L-part can be
placed in 42 different positions for each of the 8 symmetries of the
patch. 

The average branching factor is experimentally found by running 100 random
games. Both players use a simple heuristic, by choosing tiles based on
Equation~\ref{eq:gain} and placing tiles using \emph{BL-Every} and
\emph{Regret} (see Sections~\ref{sec:placing-tiles}
and~\ref{sec:experiments}), In this setting, the game averages 23.2
plies for the first player and 23 plies for the second player, with an
average branching factor of 83.2. There is also a slight first-player
advantage in this setting, with the first player winning 56 games.

\section{Placement model}
\label{sec:model}

Core for implementing a game state for Patchwork is the players
individual boards where patches are placed. Implementing placement and
packing from scratch is complicated and error-prone. We use constraint
programming to quickly and reliably implement the packing part of the
game state.

The model for placing parts on the board builds upon the placement
model for polyominoes introduced in~\cite{Lagerkvist2008}. The model
uses \cons{regular} constraints to specify the required placement of a
patch on a board. First the original model is explained briefly, and
then the additions for Patchwork are given.

\subsection{Original model}
\label{sec:model:original}

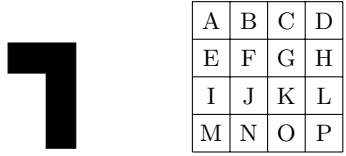
\begin{figure}
\centering
\begin{tikzpicture}[scale=0.5]
  \draw [fill] (1.1,0.1) -- (1.9,0.1) -- (1.9,2.9) -- (0.1,2.9) --
  (0.1,2.1) -- (1.1,2.1);
  \begin{scope}[xshift=5cm]
  \draw (0,0) grid (4,4);
  \node at (0.5,3.5) {A}; \node at (1.5,3.5) {B}; \node at (2.5,3.5) {C}; \node at (3.5,3.5) {D};
  \node at (0.5,2.5) {E}; \node at (1.5,2.5) {F}; \node at (2.5,2.5) {G}; \node at (3.5,2.5) {H};
  \node at (0.5,1.5) {I}; \node at (1.5,1.5) {J}; \node at (2.5,1.5) {K}; \node at (3.5,1.5) {L};
  \node at (0.5,0.5) {M}; \node at (1.5,0.5) {N}; \node at (2.5,0.5) {O}; \node at (3.5,0.5) {P};
  \end{scope}
\end{tikzpicture}
\caption{Part to place and the grid to place on.}
    \label{fig:part-and-board}
\end{figure}

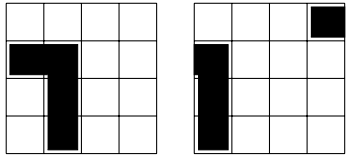
\begin{figure}
\centering
\begin{tikzpicture}[scale=0.5]
  \draw (0,0) grid (4,4);
  \draw [fill] (1.1,0.1) -- (1.9,0.1) -- (1.9,2.9) -- (0.1,2.9) -- (0.1,2.1) -- (1.1,2.1);
  
  \begin{scope}[xshift=5cm]
    \draw (0,0) grid (4,4); 
    \draw [fill] (3.1,3.1) -- (4,3.1) -- (4,3.9) -- (3.1,3.9);
    \draw [fill] (0.1,0.1) -- (0.9,0.1) -- (0.9,2.9) -- (0,2.9) --
    (0,2.1) -- (0.1,2.1);
  \end{scope}
\end{tikzpicture}
\caption{A valid (left) and an erroneous (right) placement}
    \label{fig:valid-invalid-placement}
\end{figure}

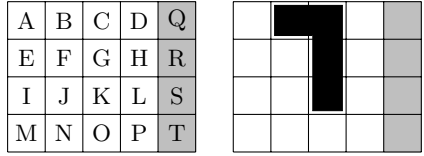
\begin{figure}[t]
\centering
\begin{tikzpicture}[scale=0.5]
  \draw [fill,lightgray] (4,0) -- (5,0) -- (5,4) -- (4,4);
  \draw (0,0) grid (5,4);
  \node at (0.5,3.5) {A}; \node at (1.5,3.5) {B}; \node at (2.5,3.5)
  {C}; \node at (3.5,3.5) {D}; \node at (4.5,3.5) {Q};
  \node at (0.5,2.5) {E}; \node at (1.5,2.5) {F}; \node at (2.5,2.5)
  {G}; \node at (3.5,2.5) {H}; \node at (4.5,2.5) {R};
  \node at (0.5,1.5) {I}; \node at (1.5,1.5) {J}; \node at (2.5,1.5)
  {K}; \node at (3.5,1.5) {L}; \node at (4.5,1.5) {S};
  \node at (0.5,0.5) {M}; \node at (1.5,0.5) {N}; \node at (2.5,0.5)
  {O}; \node at (3.5,0.5) {P}; \node at (4.5,0.5) {T};
  
  \begin{scope}[xshift=6cm]    
  \draw [fill,lightgray] (4,0) -- (5,0) -- (5,4) -- (4,4);
  \draw (0,0) grid (5,4);
  \draw [fill,xshift=1cm,yshift=1cm] (1.1,0.1) -- (1.9,0.1) -- (1.9,2.9) -- (0.1,2.9) -- (0.1,2.1) -- (1.1,2.1);
  \end{scope}

\end{tikzpicture}
  \caption{Grid extended with dummy column, and placement in new grid.}
    \label{fig:extended-board-placement}
\end{figure}

Consider the placement of the patch in Figure~\ref{fig:part-and-board}
in the 4-by-4 grid shown. Each square is represented by a
$0/1$-variable ($1$ meaning the patch covers the square), and the grid
is encoded in row-major order. All placements of the patch are encoded
by the regular expression $0^*110^310^310^*$. This encoding includes
all valid placements, but it also includes some invalid ones (see
right side in Figure~\ref{fig:valid-invalid-placement}). To forbid
such placements, an extra column of dummy squares is added that are
fixed to 0 as shown in Figure~\ref{fig:extended-board-placement}, and
the expression is changed to $0^*110^410^410^*$. Rotations and flips
are handled by combining the regular expressions for each transform
using disjunction, relying on the DFA minimization for removing states
representing equal rotations.

Unique sets $B_p$ of $0/1$ variables are used for
each patch $p$. The variables for different patches are connected
with integer variables $B$ with domain values representing patches,
empty squares, and the end column: $P\cup
\textrm{empty}\cup \textrm{end}$. While the constraint can be defined
directly on the $B$ variables, \cite{Lagerkvist2008} showed that using
the auxiliary $B_p$ variables performs much better.

\subsection{Extensions}
\label{sec:model:extensions}

The original model is not enough for implementing a model for
Patchwork. In particular, optional placements and control of
transformations is needed, as is measuring usage.

\paragraph{Transforms} For heuristics, it can be useful to branch on
which, if any, of the up to 8 transforms of a patch is placed
(transforms include rotations and flips of the patches). In order to do
this, the unique transforms are generated beforehand in the model
set-up code, instead of always generating all 8 transforms and using DFA
minimization to handle symmetries. To encode the different rotations,
8 Boolean control values $S_{ps}$ for patch $p$ in each possible transform $s$ are
prepended to each placement expression, with exactly 1 of the
variables set to 1 for each transform. Note that for simplicity the
same number of Boolean variables are prepended, regardless of the
number of non-symmetric transforms generated.

\paragraph{Optional placements} In Patchwork, not all patches are
placed on the board. While it is possible to dynamically add variables
and constraints during solving/searching, it is better to set up all
constraints for all patches from the start and controlling placement
of a patch using reification.  Given a patch $p$ with board variables $B_p$,
placement expression $R_p$ for all transforms, a $0/1$
control variable $U_p$ indicating if the patch is to be
placed, the 8 $S_p$ Boolean variables indicating which transform is
used, and a Boolean variable $N_p$ indicating that no
transform was used, the following constraint is posted:

\begin{equation}
  \label{eq:reified-placement}
  \cons{regular}((10R_p)\,|\,(010^*),\ U_pN_pS_pB_p)
\end{equation}

This is a domain-consistent full reification of the placement
constraint; when a patch can no longer be placed the propagation will
set $U_p$ to $0$. Note that the second value corresponds to no
transform chosen. This is so that the Boolean variables
$N_pS_p$ can be channeled into a single integer variable
with domain ${0..8}$

\paragraph{Usage} It is well-known that cumulative usage reasoning can
be very effective for packing problems~\cite{Simonis2008}. A first
step towards cumulative reasoning is to model resource usage. Usage
reasoning \emph{per patch} is added to the model using regular
expressions. Variables $C_{pc}^\Sigma$ for column sums and
$R_{pr}^\Sigma$ for row sums indicating the number of ones in each
column $c$ and row $r$ for patch $p$ in $B_p$. Consider again the patch in
Figure~\ref{fig:extended-board-placement}: the first column it is placed in has
a single one and the second column has three ones, while it has usage
two, one, and one for its rows. This means that $C_{p}^\Sigma$ belongs
to the language $0^*130^*$ and $R_p^\Sigma$ belongs to the language
$0^*2110^*$. Concatenating the column (including the dummy column) and
row usages into $C_p^\Sigma R_p^\Sigma$, all usages for all placements
belong to the language $0^*130^*00^*2110^*$. Using the same
infrastructure for symmetries, pre-fixing with the control variable
$U_p$ and transform $S_p$ gives a domain consistent full reification
for the usage constraint for a single patch.

\section{Placing patches on the board}
\label{sec:placing-tiles}

Armed with a model that expresses the underlying placement model,
heuristics are needed that can be used when a player chooses a patch
and needs to place it on the board. It is important to remember here
that in the context of game playing (both for MCTS-style roll-outs and
for traditional tree search), a patch needs to be fully placed,
without deciding on any of the other parts concurrently. Also
important, there is no possibility of a back-track.

We call the overall heuristic used to generate a placement on the
board a \emph{strategy}. The strategy is divided into two independent
parts, a \emph{policy} that generates placements of a tile, and an
\emph{evaluation} that chooses among different placements.  The
concept of a placement policy is common in packing literature.  The
policy is typically incomplete; most policies can not be used to find
a guaranteed optimal packing.

\subsection{Placement policy}
\label{sec:policy}

A \emph{policy} in this paper is is a method used to place a tile on a
partially filled board in some manner, potentially generating many
different placements. Policies can be implemented using standard
constraint programming branching and search. Which placement that is
used is not something the policy needs to consider. A central goal for a
policy is to cheaply produce some good placements of a part. In
constraint programming terms, a policy can be regarded as an incomplete
large-step heuristic branching without ordering of the alternatives produced.

Recall from Section~\ref{sec:model} that a patch $p$ has 0/1 variables
$B_p$ representing the placement of that patch, variables
$C_{pc}^\Sigma$ and $R_{pr}^\Sigma$ representing the number of
\mbox{occupied} squares in each column and row for the patch, and control
variables $S_p$ representing the transform of the patch.

\paragraph{Bottom-Left} One of the most common heuristics for
placement problems is the Bottom-Left heuristic~\cite{Baker1980} (BL),
in which each part is placed in the bottom-most position, and among
all bottom-most positions the left-most is chosen. Worth noting is
that in the context of a symmetric and bounded board the direction is
arbitrary (i.e., Right-Top is the same as Bottom-Left, and so on),
compared to the original context where packing was on a roll of
material that was infinite in the top/up direction.

To implement the policy, integer variables representing the first occupied
row $R_p^f$ (and $C_p^f$ for columns) are needed. For patches that are
not placed on a board the sentinel value of one past the end is
used. The row variable is defined using the following schema for the
patch $p$ using the 8 row sum variables $R_{pr}^\Sigma$ (column
variables defined similarly).

\begin{gather}
  R_p^f \in \{0..9\},\ Z_{0..8} \in \{0,1\},\ F_{0..8} \in \{0,1\} \\
  \forall_{i=0..8}\ Z_i \Leftrightarrow R_{pi}^\Sigma = 0 \\
  Z_0 = F_0,\ \forall_{i=1..8}\ F_i \Leftrightarrow F_{i-1} \land Z_i \\
  R_p^f = \sum_{i=0..8} F_i
\end{gather}

The $Z$ variables indicate if there is no placement in a row, and the
$F$ variables define if an index is part of the first run of zero
variables. Summing up $F$ gives the first index, or 9 (one past the
end) if there is no placement. This is somewhat similar to the
decomposition for \cons{length\_first\_sequence} in the Global
Constraint Catalogue~\cite{GlobalConstraintCatalogue}.

Using these variables standard constraint programming variable/value
heuristics can be used to implement Bottom-Left for the placement of
patch $p$. Branchings are added on first $C_p^f$ and then $R_p^f$
trying the smallest value first. A third branching on the $B_p$
variables is added with in-order true-first choices. Searching for the
first solution using DFS will give the Bottom-Left placement of the
patch in some transformation.

\paragraph{BL/LB} Building on the standard BL heuristic, it is easy to
extend to a heuristic that places in both Bottom-Left and Left-Bottom
order producing two results. Two individual DFS searches are made, one
with branching first on $C_p^f$ and then on $R_p^f$, and one with the
order of the branchings reversed. In both cases the $B_p$ branching is
kept the same.

\paragraph{Pareto BL} Pareto Bottom-Left is a generalization of
Bottom-Left, where instead of only finding a solution for a single
value in one direction, we try all values in that direction. This
corresponds to testing
all columns as the minimum column and finding the minimum row
placement for each. The minimum row/column values for the placements
found ($(r_i,c_i)$) form a Pareto front under the natural point wise
order relation, thus the name,

In practice, to limit the amount of branching in the
beginning and keep solutions heuristically relevant, not all columns
are tested for placement. The maximum column with some fixed placement
on the board $c_{\max}$ is found before search, and only columns
in $0..c_{\max}+1$ are used. For a partial packing that is still close
to a corner, this means that non-needed placements close to the other
corner are not tested.

Given a set of columns to test, the implementation manually assigns
the $C_p^f$ to each values in $0..c_{\max}+1$. After this step, for
each column the search proceeds as for Bottom-Left.

\paragraph{Policies based on standard heuristics} There is a rich area
of research into general heuristics for constraint programming. We use
five different general heuristics that are available in Gecode, the
system used for implementation. The heuristic values used are:

\begin{description}
\item[In Order] Use the order of the variables. This is actually quite
  close to a Bottom-Left style heuristic, but instead of working on
  the bounding box of the patch, it works on the individual squares.
\item[Size] The domain size for variables. Commonly called
  \emph{First fail}.
\item[AFC] The accumulated failure count~\cite{Boussemart04}
  (also known as \emph{weighted degree}) is the sum of all the
  times propagators connected to the variable have failed a search
  tree.
\item[Action] Action is the number of times the domain of a variable
  has been reduced~\cite{Michel12} (also known as \emph{activity}).
\item[CHB] CHB (Conflict History-based Branching,
  \cite{Liang16,MPG:6.2.0}) uses a combination of domain reduction
  counts and when failures occur.
\end{description}

The branching is done on the $B_p$ Boolean variables, but for all but
the simplest (\emph{In Order}), the heuristic values would not make
much sense on the $B_p$ variables: the domain size is always 2 and
there is typically not much activity on the Boolean variables. The
base version is to use the heuristic value associated with the
corresponding $B$ variable that represents all the patches. For
\emph{AFC}, \emph{Activity}, and \emph{CHB} variants are also defined
that sum the heuristic value over both the corresponding $B$ variable
and the corresponding $B_t$ variables for \emph{all} patches $t$ (including
the current patch $p$). The summed variants are called
\emph{$\sum$AFC}, \emph{$\sum$Activity}, and \emph{$\sum$CHB}.

For \emph{AFC}, \emph{Activity}, \emph{CHB}, and their summed
variants, the heuristic values that are used in the policies are
divided by the domain size of the variable, as is common. This is
denoted \emph{X/Size} for the measure \emph{X}.

For all the standard heuristics, equal values use the order of the
variables as a tie-breaker.

\paragraph{Every transform} The above heuristics all produce a result in
some transformation, without actually making any specific choice. For all
the heuristics, variants that create placements for all possible
rotations of a patch are also tested.

To implement an \emph{every transform} variant for a placement policy,
all possible values for the rotation variable $U_p$ are assigned
first. For each assignment, the base policy is applied to generate a
placement of the patch.

\paragraph{All} Finally, just generating all the possible placement
and letting the evaluation make the choice is possible. This is simply
implemented as a standard DFS search for all solutions when branching
over the $B_p$ variables. Note that \emph{All} will naturally always
generate solutions for all rotations, so it does not make sens to use
the \emph{every transform} modification.

\subsection{Placement evaluation}
\label{sec:placement-evaluation}

Many of the policies produce more than one alternative. To choose
among the various alternatives, an \emph{evaluation} is used that
takes multiple placements and returns the ``best'' one. In constraint
programming terms, an evaluation can be seen as an ordering of the
alternatives produced by a branching, so that a left-most exploration
order would follow the evaluations heuristic. Compared to many
constraint programming branchings, the evaluation expects the
alternatives to be of equal type; they would not work for classical
$c\lor \lnot c$ two-way branching such as $x=d\lor x\neq d$.

Each placement evaluated by the evaluation is represented by the full
search state with the patch placed and all the other board variables
present and all constraints fully propagated. Also, the evaluations
have access to the state before the placement, for comparisons.

\paragraph{First and Random} The evaluation \emph{First}  simply
chooses the first alternative. This is not a meaningful evaluation
(except for heuristics that only generate one placement), but is
interesting as a baseline. Similarly, the \emph{Random} evaluation
chooses one of the alternatives at random.

\paragraph{Bottom. Left, and Area} A placement of a patch has a maximum
extent to the right and to the top. The \emph{Left} evaluation chooses
the placement with minimum right extent, while the \emph{Bottom}
evaluation chooses the minimum top extent. The \emph{Area} evaluation
measures the increase in the bounding box of the fixed placements,
promoting placements that are kept tight to a corner.

\paragraph{Propagation Guided Global Regret} When placing a patch,
propagation will remove possible placements for other patches. This is
a valuable signal on how many possibilities we have left, and can be
used to guide the search. 

We call this the \emph{Propagation Guided Global Regret}, since it
uses propagation to give an indication of the effect of a choice
globally. More formally, we define it as follows. Let the original
variables for the whole board be $B$, and the variables after a
placement and propagation of that placement be $B'$. The expression
$B_{ij}$ represents the square at indexes $i$ and $j$, and $|B_{ij}|$
represents the domain size of the variable. The patch to place is
named $p$. The heuristic value is defined by the following summation.

\begin{equation}
  \label{eq:regret}
  \mathrm{pggr}(B, B', p) = \sum_{i=0}^8\sum_{j=0}^8
  \begin{cases}
    0 & \text{if } |B_{ij}| = 1 \\
    0 & \text{if } B'_{ij} = p \\
    |B_{ij}| - |B'_{ij}| & \text{otherwise}
  \end{cases}
\end{equation}

Regret is a well-known heuristic in constraint programming, and is
usually defined as the difference between the minimum and next to
minimum (and similarly for maximum) value in a domain. This is mostly
useful when there is a direct connection between variable domain
values and some optimization criteria. However, the concept of regret
is more general than the typical single-variable domain centric value optimizing
view. Here we lift the concept to be with respect to keeping possible
future options open.

In a sense, propagation guided global regret is the anti-thesis of
Impact based search~\cite{Refalo04}, where variable/value pairs are
chosen based on the amount of propagation that they trigger
historically. The context here is very different though, with the
amount of propagation used is a negative signal. In addition, it is
used for the current assignment using look-ahead, and not based on
statistics of historic values.

\section{Experimental evaluation}
\label{sec:experiments}

This section reports results of experimental evaluation to clarify how
the different strategies formed by combinations of the policies and
evaluations introduced perform.

\subsection{Implementation and Execution Environment}
\label{sec:experiments:implementation}

The implementation consists of a game state implemented in C++17 with
the placement model implemented using the Gecode~\cite{gecode}
constraint programming system version 6.2.0. The code is
single-threaded only. While it would be possible to use multi-threaded
search, single-threaded execution gives a simple and more stable
evaluation.  The implementation is available at
\url{htts://github.com/zayenz/cp-mod-ref-2019-patchwork}. All
experiments are run on a Macbook Pro 15 with a 6-core 2.7 GHz Intel
Core i7 processor and 16 GiB memory.

For the learning heuristics (those that use \emph{AFC}, \emph{Activity},
and \emph{CHB}) the statistics they are based on are collected
throughout the whole experiment, giving them their best possible chance
of learning interesting aspects.

\subsection{Core packing problem}
\label{sec:experiments:packing}

The central aspect in the game play is that a sequence of patches are
chosen and placed on the board. Each placement must be done without
knowing which future patches will be placed. A pure packing problem
that captures this is to order all the patches, and to test placing
each patch in turn, incrementally building up a patchwork.

In table~\ref{tab:placement} results are shown for testing the same 1000 random
orders of parts on 119 strategies. The strategies are combinations of
policies with \emph{Some} or \emph{Every} transformation matched with
different evaluations. Each strategy has four metrics. For each
metric, the best value is marked with dark blue, those within 1\% of
the best value are marked with medium blue, and those within 5\% are
marked with light blue background. The metrics are

\begin{description}
\item[Area] The mean amount of area placed after all patches have been
  placed. This is the main metric in the game, and is a natural metric
  on the power of a strategy to make good placements without knowing
  what patches to try next.
\item[Streak] The mean number of patches placed before the
  first failure. This is a measure of how much the strategy manages to
  keep options open, but does not necessarily correspond to the amount
  of area placed.
\item[Time] The mean time spent making a single placement of a patch
  in milliseconds.
\item[Alts] The mean number of alternatives produced for each patch.
\end{description}

Note that for strategies that only create one alternative, the
evaluation does not matter, but is tested for completeness. The
evaluation \emph{ReverseRegret} (choosing the maximum instead of
minimum regret) is added to show the behaviour when
explicitly going against the intuition that keeping options open for
packing is a good idea.

For all learning heuristics, only the version with the heuristic value
divided by domain size is tested. We do not show the results for
\emph{$\sum$AFC}, \emph{$\sum$Activity}, and \emph{CHB} since they
were slightly worse than their corresponding summed/non-summed
variants. The full data is available on request.

\begin{sidewaystable}
  \centering
  \tiny
\begin{tabular}{lll@{\hspace{1em}}rr@{\hspace{1em}}rr@{\hspace{1em}}rr@{\hspace{1em}}rr@{\hspace{1em}}rr@{\hspace{1em}}rr@{\hspace{1em}}rr@{\hspace{1em}}rr@{\hspace{1em}}rl}
\toprule
 & &  & \multicolumn{2}{c}{In Order}  & \multicolumn{2}{c}{Size}  & \multicolumn{2}{c}{AFC/Size}  & \multicolumn{2}{c}{Action/Size}  & \multicolumn{2}{c}{$\sum$CHB/Size}  & \multicolumn{2}{c}{BL}  & \multicolumn{2}{c}{BL/LB}  & \multicolumn{2}{c}{Pareto BL}  & \multicolumn{1}{c}{All} & \\
         & &  & \multicolumn{1}{c}{Some}  & \multicolumn{1}{c}{Every}  & \multicolumn{1}{c}{Some}  & \multicolumn{1}{c}{Every}  & \multicolumn{1}{c}{Some}  & \multicolumn{1}{c}{Every}  & \multicolumn{1}{c}{Some}  & \multicolumn{1}{c}{Every}  & \multicolumn{1}{c}{Some}  & \multicolumn{1}{c}{Every}  & \multicolumn{1}{c}{Some}  & \multicolumn{1}{c}{Every}  & \multicolumn{1}{c}{Some}  & \multicolumn{1}{c}{Every}  & \multicolumn{1}{c}{Some}  & \multicolumn{1}{c}{Every}  & \multicolumn{1}{c}{Every} & \\
\midrule
 & \multirow{4}{*}{First} & \whitebox{Area}  & \closetobest{76.51} & \whitebox{72.25} & \whitebox{66.49} & \whitebox{65.95} & \whitebox{73.73} & \whitebox{71.68} & \closetobest{74.97} & \whitebox{71.70} & \closetobest{74.98} & \whitebox{72.16} & \closetobest{76.12} & \whitebox{72.28} & \closetobest{76.12} & \whitebox{72.28} & \closetobest{76.12} & \whitebox{73.75} & \closetobest{76.51}\\
 &  & \whitebox{Streak}  & \whitebox{15.79} & \whitebox{14.06} & \whitebox{12.22} & \whitebox{12.50} & \whitebox{14.50} & \whitebox{13.72} & \whitebox{15.52} & \whitebox{14.15} & \whitebox{14.85} & \whitebox{13.72} & \whitebox{15.62} & \whitebox{14.46} & \whitebox{15.62} & \whitebox{14.46} & \whitebox{15.62} & \whitebox{14.88} & \whitebox{15.79}\\
 &  & \whitebox{Time}  & \whitebox{25.37} & \whitebox{86.90} & \almostbest{24.22} & \whitebox{83.72} & \closetobest{24.72} & \whitebox{87.86} & \closetobest{25.16} & \whitebox{88.90} & \closetobest{25.01} & \whitebox{88.90} & \whitebox{162.30} & \whitebox{316.35} & \whitebox{217.67} & \whitebox{501.64} & \whitebox{722.31} & \whitebox{1669.19} & \whitebox{1815.18}\\
 &  & \whitebox{Alts}  & \whitebox{1.00} & \whitebox{2.99} & \whitebox{1.00} & \whitebox{2.95} & \whitebox{1.00} & \whitebox{2.97} & \whitebox{1.00} & \whitebox{3.03} & \whitebox{1.00} & \whitebox{2.97} & \whitebox{1.00} & \whitebox{3.04} & \whitebox{1.94} & \whitebox{6.02} & \whitebox{3.64} & \whitebox{9.62} & \best{71.86}\\
& \\ & \multirow{4}{*}{Random} & \whitebox{Area}  & \closetobest{76.51} & \whitebox{72.25} & \whitebox{66.49} & \whitebox{66.18} & \whitebox{73.73} & \whitebox{71.63} & \closetobest{74.97} & \whitebox{71.50} & \closetobest{74.98} & \whitebox{72.08} & \closetobest{76.12} & \whitebox{72.35} & \closetobest{75.43} & \whitebox{71.91} & \whitebox{72.72} & \whitebox{70.62} & \whitebox{69.49}\\
 &  & \whitebox{Streak}  & \whitebox{15.79} & \whitebox{14.26} & \whitebox{12.22} & \whitebox{12.27} & \whitebox{14.50} & \whitebox{13.60} & \whitebox{15.51} & \whitebox{14.02} & \whitebox{14.85} & \whitebox{13.71} & \whitebox{15.62} & \whitebox{14.36} & \whitebox{15.66} & \whitebox{14.18} & \whitebox{14.19} & \whitebox{13.10} & \whitebox{12.29}\\
 &  & \whitebox{Time}  & \whitebox{26.03} & \whitebox{89.09} & \almostbest{24.15} & \whitebox{82.85} & \closetobest{24.76} & \whitebox{87.68} & \closetobest{25.11} & \whitebox{88.47} & \closetobest{25.05} & \whitebox{88.42} & \whitebox{161.86} & \whitebox{317.08} & \whitebox{210.55} & \whitebox{497.82} & \whitebox{1208.40} & \whitebox{2793.32} & \whitebox{1197.45}\\
 &  & \whitebox{Alts}  & \whitebox{1.00} & \whitebox{3.02} & \whitebox{1.00} & \whitebox{2.91} & \whitebox{1.00} & \whitebox{2.98} & \whitebox{1.00} & \whitebox{3.02} & \whitebox{1.00} & \whitebox{2.96} & \whitebox{1.00} & \whitebox{3.02} & \whitebox{1.94} & \whitebox{5.99} & \whitebox{4.87} & \whitebox{11.22} & \whitebox{51.41}\\
& \\ & \multirow{4}{*}{Left} & \whitebox{Area}  & \closetobest{76.51} & \whitebox{72.79} & \whitebox{66.49} & \whitebox{66.05} & \whitebox{73.73} & \whitebox{72.03} & \closetobest{74.99} & \whitebox{71.91} & \closetobest{74.97} & \whitebox{72.48} & \closetobest{76.12} & \whitebox{73.47} & \closetobest{76.11} & \whitebox{73.47} & \closetobest{75.93} & \whitebox{74.47} & \closetobest{75.08}\\
 &  & \whitebox{Streak}  & \whitebox{15.79} & \whitebox{14.08} & \whitebox{12.22} & \whitebox{12.18} & \whitebox{14.50} & \whitebox{13.70} & \whitebox{15.52} & \whitebox{14.22} & \whitebox{14.85} & \whitebox{13.97} & \whitebox{15.62} & \whitebox{14.91} & \whitebox{15.62} & \whitebox{14.91} & \whitebox{15.51} & \whitebox{15.13} & \whitebox{14.95}\\
 &  & \whitebox{Time}  & \closetobest{25.24} & \whitebox{88.13} & \best{24.13} & \whitebox{82.60} & \closetobest{24.75} & \whitebox{88.46} & \closetobest{25.11} & \whitebox{89.28} & \closetobest{24.98} & \whitebox{89.49} & \whitebox{162.11} & \whitebox{303.51} & \whitebox{217.95} & \whitebox{447.45} & \whitebox{726.92} & \whitebox{1127.38} & \whitebox{1657.31}\\
 &  & \whitebox{Alts}  & \whitebox{1.00} & \whitebox{3.00} & \whitebox{1.00} & \whitebox{2.91} & \whitebox{1.00} & \whitebox{2.98} & \whitebox{1.00} & \whitebox{3.03} & \whitebox{1.00} & \whitebox{2.97} & \whitebox{1.00} & \whitebox{3.05} & \whitebox{1.94} & \whitebox{6.05} & \whitebox{3.60} & \whitebox{7.95} & \whitebox{65.69}\\
& \\ & \multirow{4}{*}{Bottom} & \whitebox{Area}  & \closetobest{76.51} & \whitebox{72.25} & \whitebox{66.49} & \whitebox{65.95} & \whitebox{73.73} & \whitebox{71.68} & \closetobest{75.00} & \whitebox{71.74} & \closetobest{74.98} & \whitebox{72.16} & \closetobest{76.12} & \whitebox{72.28} & \closetobest{76.12} & \whitebox{72.28} & \closetobest{76.12} & \whitebox{73.75} & \closetobest{76.51}\\
 &  & \whitebox{Streak}  & \whitebox{15.79} & \whitebox{14.06} & \whitebox{12.22} & \whitebox{12.50} & \whitebox{14.50} & \whitebox{13.72} & \whitebox{15.54} & \whitebox{14.15} & \whitebox{14.85} & \whitebox{13.72} & \whitebox{15.62} & \whitebox{14.46} & \whitebox{15.62} & \whitebox{14.46} & \whitebox{15.62} & \whitebox{14.88} & \whitebox{15.79}\\
 &  & \whitebox{Time}  & \closetobest{25.30} & \whitebox{86.58} & \almostbest{24.13} & \whitebox{83.87} & \closetobest{24.67} & \whitebox{87.80} & \closetobest{25.13} & \whitebox{89.22} & \closetobest{25.03} & \whitebox{88.71} & \whitebox{162.03} & \whitebox{315.70} & \whitebox{217.42} & \whitebox{501.39} & \whitebox{722.04} & \whitebox{1668.19} & \whitebox{1815.77}\\
 &  & \whitebox{Alts}  & \whitebox{1.00} & \whitebox{2.99} & \whitebox{1.00} & \whitebox{2.95} & \whitebox{1.00} & \whitebox{2.97} & \whitebox{1.00} & \whitebox{3.03} & \whitebox{1.00} & \whitebox{2.97} & \whitebox{1.00} & \whitebox{3.04} & \whitebox{1.94} & \whitebox{6.02} & \whitebox{3.64} & \whitebox{9.62} & \best{71.86}\\
& \\ & \multirow{4}{*}{Area} & \whitebox{Area}  & \closetobest{76.51} & \whitebox{72.79} & \whitebox{66.49} & \whitebox{66.05} & \whitebox{73.73} & \whitebox{72.03} & \closetobest{74.98} & \whitebox{71.91} & \closetobest{74.98} & \whitebox{72.48} & \closetobest{76.12} & \whitebox{73.47} & \closetobest{76.11} & \whitebox{73.47} & \closetobest{75.93} & \whitebox{74.47} & \closetobest{75.08}\\
 &  & \whitebox{Streak}  & \whitebox{15.79} & \whitebox{14.08} & \whitebox{12.22} & \whitebox{12.18} & \whitebox{14.50} & \whitebox{13.70} & \whitebox{15.51} & \whitebox{14.24} & \whitebox{14.85} & \whitebox{13.97} & \whitebox{15.62} & \whitebox{14.91} & \whitebox{15.62} & \whitebox{14.91} & \whitebox{15.51} & \whitebox{15.13} & \whitebox{14.95}\\
 &  & \whitebox{Time}  & \closetobest{25.26} & \whitebox{88.03} & \almostbest{24.16} & \whitebox{82.82} & \closetobest{24.67} & \whitebox{88.14} & \closetobest{25.12} & \whitebox{89.41} & \closetobest{25.07} & \whitebox{89.45} & \whitebox{161.79} & \whitebox{303.60} & \whitebox{217.66} & \whitebox{447.02} & \whitebox{727.14} & \whitebox{1129.46} & \whitebox{1652.75}\\
 &  & \whitebox{Alts}  & \whitebox{1.00} & \whitebox{3.00} & \whitebox{1.00} & \whitebox{2.91} & \whitebox{1.00} & \whitebox{2.98} & \whitebox{1.00} & \whitebox{3.03} & \whitebox{1.00} & \whitebox{2.97} & \whitebox{1.00} & \whitebox{3.05} & \whitebox{1.94} & \whitebox{6.05} & \whitebox{3.60} & \whitebox{7.95} & \whitebox{65.69}\\
& \\ & \multirow{4}{*}{Regret} & \whitebox{Area}  & \closetobest{76.51} & \closetobest{77.13} & \whitebox{66.49} & \whitebox{69.67} & \whitebox{73.73} & \closetobest{75.11} & \closetobest{75.00} & \closetobest{76.23} & \closetobest{74.98} & \closetobest{75.75} & \closetobest{76.12} & \closetobest{77.20} & \closetobest{77.21} & \almostbest{78.21} & \almostbest{77.86} & \almostbest{78.36} & \best{78.40}\\
 &  & \whitebox{Streak}  & \whitebox{15.79} & \closetobest{16.21} & \whitebox{12.22} & \whitebox{13.76} & \whitebox{14.50} & \whitebox{15.36} & \whitebox{15.56} & \closetobest{15.99} & \whitebox{14.85} & \whitebox{15.52} & \whitebox{15.62} & \closetobest{16.24} & \closetobest{16.39} & \best{16.76} & \closetobest{16.43} & \almostbest{16.65} & \closetobest{16.56}\\
 &  & \whitebox{Time}  & \closetobest{25.29} & \whitebox{96.33} & \almostbest{24.23} & \whitebox{89.96} & \closetobest{24.71} & \whitebox{95.29} & \closetobest{25.06} & \whitebox{98.15} & \closetobest{25.03} & \whitebox{96.50} & \whitebox{161.95} & \whitebox{355.38} & \whitebox{223.62} & \whitebox{606.60} & \whitebox{1426.38} & \whitebox{3550.62} & \whitebox{1464.55}\\
 &  & \whitebox{Alts}  & \whitebox{1.00} & \whitebox{3.17} & \whitebox{1.00} & \whitebox{3.03} & \whitebox{1.00} & \whitebox{3.12} & \whitebox{1.00} & \whitebox{3.21} & \whitebox{1.00} & \whitebox{3.11} & \whitebox{1.00} & \whitebox{3.17} & \whitebox{1.94} & \whitebox{6.46} & \whitebox{4.76} & \whitebox{12.39} & \whitebox{59.48}\\
& \\ & \multirow{4}{*}{ReverseRegret} & \whitebox{Area}  & \closetobest{76.51} & \whitebox{68.91} & \whitebox{66.49} & \whitebox{63.34} & \closetobest{75.14} & \whitebox{68.97} & \whitebox{74.12} & \whitebox{68.54} & \closetobest{74.78} & \whitebox{69.69} & \closetobest{76.12} & \whitebox{68.75} & \whitebox{73.34} & \whitebox{67.57} & \whitebox{67.46} & \whitebox{64.70} & \whitebox{62.73}\\
 &  & \whitebox{Streak}  & \whitebox{15.79} & \whitebox{12.82} & \whitebox{12.22} & \whitebox{11.03} & \whitebox{15.46} & \whitebox{12.68} & \whitebox{14.90} & \whitebox{12.39} & \whitebox{14.96} & \whitebox{12.34} & \whitebox{15.62} & \whitebox{12.80} & \whitebox{14.93} & \whitebox{12.32} & \whitebox{11.59} & \whitebox{10.57} & \whitebox{9.49}\\
 &  & \whitebox{Time}  & \whitebox{26.64} & \whitebox{84.63} & \closetobest{24.81} & \whitebox{79.56} & \whitebox{26.24} & \whitebox{84.72} & \whitebox{25.72} & \whitebox{83.81} & \whitebox{25.58} & \whitebox{83.63} & \whitebox{164.97} & \whitebox{278.77} & \whitebox{187.41} & \whitebox{404.47} & \whitebox{1174.86} & \whitebox{2116.79} & \whitebox{1195.69}\\
 &  & \whitebox{Alts}  & \whitebox{1.00} & \whitebox{2.92} & \whitebox{1.00} & \whitebox{2.82} & \whitebox{1.00} & \whitebox{2.93} & \whitebox{1.00} & \whitebox{2.90} & \whitebox{1.00} & \whitebox{2.87} & \whitebox{1.00} & \whitebox{2.92} & \whitebox{1.94} & \whitebox{5.72} & \whitebox{4.94} & \whitebox{9.82} & \whitebox{52.05}\\

\bottomrule
\end{tabular}

  \caption{Results for combinations of policy and evaluation
    for the core packing problem. Best values for each metric are
    indicated with blue, darker being better.}
  \label{tab:placement}
\end{sidewaystable}

\subsection{Results of packing experiments}
\label{sec:experiments:results}

The most striking observation, is that using propagation guided global
regret is clearly the most important factor in maximizing the
amount of area placed. The fact that regret is best when trying all
placements is a strong indication that it is an evaluation that can
truly guide the placement, and not just choose among a set of probably
good placements. The reverse of regret is clearly the worst among all
evaluations, validating the assumption that regret is a good measure.

Among the other evaluations, it is worth noting that when using a
heuristic that places first bottom then left, it is better to use an
evaluation that agrees with the direction placements are made in (that
is, for Bottom-Left it is better to make choices based on bottomness
than leftness).

Given a reasonable evaluation (that is, not \emph{First},
\emph{Random}, nor \emph{ReverseRegret}), the policies based on
packing heuristics are better than generic constraint programming
heuristics. The most interesting stand-out here is \emph{In Order},
since it combines a very simple constraint programming definition and
speed of evaluation with actually being similar in effect to classical
packing heuristics.

In implementing game tree search, the time used for expanding nodes is
crucial to get reasonable performance, since there is typically a hard
limit on the deliberation time of an agent. It is clear that many of
the strategies take much more time than would be feasible. Some simple
profiling of the code indicates that there is some overhead in the
implementation in how many clones are generated of the Gecode search
spaces, which could potentially be optimized. Parallel execution is
also a clear potential for improving time.

In game play, combinations of strategies can be used. For example, when
implementing MCTS it is common to use different strategies for the
tree expansion and for the roll outs. For tree expansion, quality of
moves and strong ordering is important, while roll-outs have very
strong speed requirements. A combination of a more expensive tree
expansion such as \emph{BL/LB-Every + Regret}, and a simple and fast
roll-out strategy such as \emph{In Order-Some + First} could be
useful.

In conclusion, the most important decision is to use propagation guided
global regret to choose among possible placements, with the choice of
policy guided by the time-requirements needed.

\section{Conclusions}
\label{sec:conclusions}

This paper has introduced the use of constraint programming for
representing parts of a game state for the game Patchwork. The model
represents a packing of an unknown subset of polyominoes, that are to
be chosen during game tree search.  The packing model extended and
improved a previous polyomino packing model based on \cons{regular}
constraints. The use of constraint programming simplified the task of
implementing the packing part of the game state, with a high-level
specification.

To guide the search, several new strategies were developed for placing
patches. Placement policies inspired by classical packing literature
are shown to be good. For choosing among different placements, the
concept of propagation guided global regret was introduced and shown
to be very effective in guiding search towards good placements of
patches.

\paragraph{Future work} This paper has focused on the strategies used
for placing polyominoes when implementing the  game
Patchwork. The natural next step is to also apply the model in 
game playing agents. An investigation into the relative complexity of
implementing the packing model without the support of a constraint
programming system would also be interesting. In particular, something
like global propagation guided regret would be very hard to implement
by hand.

\section*{Acknowledgments}
Thanks to Magnus Gedda, Magnus Rattfeldt, and Martin Butler for
interesting and informative discussions on games and search methods.
Thank the anonymous reviewers for their comments that helped improve
this paper.

\bibliographystyle{splncs04}
\bibliography{references}

\end{document}